%% file: main.tex
\documentclass[10pt,twocolumn,letterpaper]{article}

\usepackage{wacv}              


\definecolor{wacvblue}{rgb}{0.21,0.49,0.74}
\usepackage[pagebackref,breaklinks,colorlinks,allcolors=wacvblue]{hyperref}
\usepackage{hyperref}

\def\wacvPaperID{2350} 
\def\confName{WACV}
\def\confYear{2027}

\title{From Adults to Children: Benchmarking Foundation Models for Chest X-Ray Classification under Domain Shift}

\author{
Sara Inácio$^{1,3}$, Hugo Proença$^{1,2}$, João C. Neves$^{1,3}$\\[4pt]
\small $^{1}$University of Beira Interior, Portugal \quad \small $^{2}$IT: Instituto de Telecomunicações \small \quad $^{3}$NOVA LINCS\\
\small \texttt{sara.inacio@ubi.pt, hugomcp@ubi.pt, jcneves@ubi.pt}
}

\usepackage[table]{xcolor}
\usepackage{multirow}
\usepackage{makecell}
\usepackage{graphicx}
\usepackage{array}
\usepackage[normalem]{ulem}
\newcommand{\std}[1]{{\scriptsize $\pm$ #1}}
\useunder{\uline}{\ul}{}

\definecolor{benchmarkblue}{HTML}{24456B}
\definecolor{paramblue}{HTML}{D9E2F3}
\definecolor{domaingreen}{HTML}{C6E0B4}
\definecolor{intraorange}{HTML}{F8CBAD}
\definecolor{sectiongreen}{HTML}{A9D18E}

\usepackage[table]{xcolor}
\usepackage{multirow}
\usepackage{makecell}

\definecolor{headerblue}{HTML}{D9E2F3}
\definecolor{domaingreen}{HTML}{C6E0B4}
\definecolor{intraorange}{HTML}{F8CBAD}
\definecolor{sectiongreen}{HTML}{A9D18E}

\begin{document}
\maketitle
\input{0_abstract}    
\input{1_intro}
\input{2_relatedwork}

\input{3_benchmark}
\input{4_evaluation}
\input{5_results}
\input{6_discussion}
\input{7_limitations}
\input{8_conclusion}
{
    \small
    \bibliographystyle{ieeenat_fullname}
    \bibliography{main}
}

\end{document}

%% file: 0_abstract.tex
\begin{abstract}
Foundation models have shown significant potential across diverse fields and have been applied to medical imaging, but their robustness across patient populations remains unclear. In this work, we investigate foundation models' ability to generalize from adult to pediatric chest X-ray classification. To the best of our knowledge, this is the first benchmark to compare multiple foundation models under a common adult-to-pediatric evaluation setting while also analyzing the effect of adaptation depth. We harmonize VinDr-CXR and PediCXR into a five-label classification task and evaluate six foundation models and two CNN baselines. All models are adapted using adult data only and tested in both the in-domain and pediatric domains using linear probing, partial fine-tuning, and full fine-tuning. The benchmark reveals a clear and consistent trend: all models suffer a substantial performance degradation when transferred from adult to pediatric data, with drops exceeding 10 percentage points. While Macro-AUROC values reach approximately 92\% on the adult test set, performance decreases to around 78\% on the pediatric domain. Another important finding is that increasing the depth of adaptation improves pediatric performance on average, although the effect varies across models. Moreover, strong in-domain performance does not necessarily indicate greater robustness to domain shift.
\vspace{-0.2cm}

\end{abstract}
\vspace{-0.4cm}

%% file: 1_intro.tex
\section{Introduction}
\label{sec:intro}

\begin{figure}[t]
    \centering
    \includegraphics[width=1\linewidth]{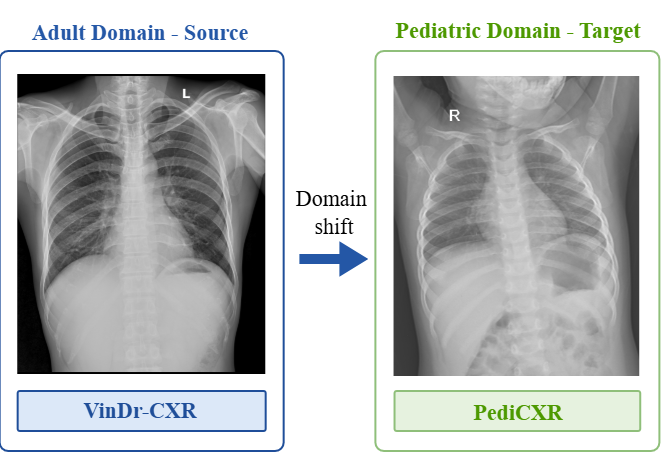}
    \caption{\textbf{Illustration of the domain shift between adult and pediatric chest X-rays.} The model is trained on the adult source domain and evaluated on the pediatric target domain.}
    \label{fig:placeholder}
    \vspace{-0.5cm}
\end{figure}

\vspace{-0.1cm}
Medical imaging plays an important role in clinical diagnosis, and deep learning has shown considerable potential for automated medical image analysis. Chest radiography (CXR) is widely used in pediatric care due to its low cost, availability, and relatively low radiation exposure~\cite{padash2022pediatric}. However, the development of deep learning systems for chest X-ray analysis has been largely supported by adult datasets. Previous studies have shown that models trained on adult chest radiographs can experience substantial performance degradation when evaluated on pediatric populations~\cite{chatterjee2025children, cheng2024self, zunaed2024improving}. As illustrated in Fig. 1, this problem is closely related to domain shift~\cite{moreno2012unifying}, where the data distribution at test time differs from that observed during training. In medical imaging, such differences may arise from acquisition devices, imaging protocols, clinical institutions, annotation practices, or patient populations~\cite{guan2021domain}. Age is a particularly relevant factor in population shifts in chest radiography, as adult and pediatric examinations differ in anatomical characteristics, disease prevalence, and the presentation of several radiographic findings. Foundation models \cite{bommasani2021opportunities} have recently emerged as a promising approach for learning transferable representations from large-scale datasets. In medical imaging, these models are commonly pretrained using self-supervised, weakly supervised, or multimodal objectives and can then be adapted to downstream clinical tasks \cite{van2025foundation}. Several foundation models have also been developed specifically for medical imaging and chest radiography \cite{zhang2023biomedclip, bannur2023biovilt, perez2025rad-dino}. Their transfer capabilities make them particularly relevant for studying whether large-scale pretraining can improve robustness across patient populations. Despite this progress, the behavior of foundation models under an adult-to-pediatric CXR domain shift remains poorly characterized. Existing work has mainly focused on individual models, specific tasks, or single adaptation settings. In particular, it remains unclear how foundation models with different pretraining domains and learning objectives compare when evaluated under the same adult-to-pediatric protocol, and how the level of downstream adaptation affects their cross-domain performance. In this work, we study adult-to-pediatric generalization in chest X-ray classification by conducting a standardized comparison between conventional CNNs and foundation models. We evaluate two CNN baselines and six foundation models spanning different pretraining domains and learning objectives. All models are adapted exclusively using adult VinDr-CXR \cite{vindr} data and evaluated both in-domain on adult radiographs and cross-domain on PediCXR \cite{pedicxr}, without using pediatric data during downstream adaptation. We evaluate each model under three adaptation strategies: linear probing, partial fine-tuning, and full fine-tuning. To the best of our knowledge, this is the first systematic benchmark of multiple foundation models under a common adult-to-pediatric CXR evaluation protocol. Our main contributions are:
\begin{itemize}

    \item We provide a standardized benchmark of conventional CNNs and foundation models under an adult-to-pediatric chest X-ray domain shift.
    
    \item We evaluate three downstream adaptation strategies: linear probing, partial fine-tuning, and full fine-tuning, and study how they affect in-domain and cross-domain performance.
    
    \item We analyze how model family, pretraining domain, and learning objective relate to the performance degradation observed between adult and pediatric populations.
\end{itemize}

\noindent The benchmark repository is available at {\small\url{https://github.com/sarainacio/Adult-to-Pediatric-CXR-Benchmark}}.
\vspace{-0.1cm}


%% file: 2_relatedwork.tex
\section{Related Work}
\label{sec:related_work}

\subsection{Foundation Models for Medical Imaging}
Foundation models have become increasingly adopted in medical imaging, offering an alternative to task-specific supervised representation learning. Early work in this area focused on vision-language pretraining, following the success of models such as CLIP \cite{radford2021learning} in learning transferable representations from large collections of image–text pairs. This idea translated to medical data, where images are often accompanied by free-text reports, and showed that radiology reports can provide useful semantic supervision and support transfer to downstream imaging tasks without relying on extensive manual annotation. More recent methods have explored stronger forms of domain specialization. MedCLIP \cite{wang2022medclip} adapts contrastive vision-language learning to medical data while reducing the dependence on strictly paired image–text samples, whereas BioMedCLIP \cite{zhang2023biomedclip} expands the scale of biomedical pretraining using millions of image–text pairs drawn from the scientific literature. At the same time, image-only foundation models have shown that textual supervision is not a prerequisite for transferable visual representations. DINOv2 \cite{oquab2023dinov2}, despite being trained on general-domain images, achieves strong transfer across visual tasks through self-supervised learning. These models therefore provide a useful basis for studying how architecture, pretraining modality, and domain specificity affect transfer to medical imaging tasks.
\vspace{-0.2cm}
\subsection{Foundation Models for Chest X-rays}
Chest radiography has become an important application area for foundation models because several large datasets provide radiographs together with their corresponding reports. This has made vision-language pretraining particularly attractive. BioViL \cite{boecking2022biovil}, for instance, learns radiology-specific representations from paired chest X-rays and reports, and BioViL-T \cite{bannur2023biovilt} extends this idea to longitudinal studies by incorporating temporal information. CheXzero \cite{tiu2022chexzero} and CXR-CLIP \cite{you2023cxr-clip} also rely on image–text contrastive learning, with a focus on zero-shot and transferable recognition. ELIXR \cite{xu2023elixr} takes a somewhat different direction, linking chest radiograph representations with large language models. Image-only pretraining has also received increasing attention in chest radiography. RAD-DINO \cite{perez2025rad-dino} applies self-supervised DINO-style training to large chest X-ray collections, allowing visual representations to be learned without using the associated reports. More recent models such as RayDINO \cite{moutakanni2024raydino} and EVA-X \cite{yao2025evax} follow a similar direction, further exploring large-scale self-supervised learning for radiographic images.
\vspace{-0.1cm}
\subsection{Domain Shift in Medical Imaging}
In many deep learning settings, it is implicitly assumed that the training and test data follow the same distribution, or at least a very similar one. Medical images are produced in complex clinical environments, and their appearance is affected by several factors \cite{guan2021domain}. Differences in scanners, acquisition protocols, institutions, patient populations, and data collection periods can lead to variations between datasets, even when they address the same clinical problem. As a result, a model that performs well on data from one domain may show a considerable drop in performance when evaluated on another domain. Previous studies have reported this effect across different sources of domain shift in medical imaging, including acquisition-related differences, changes between clinical sites, and variations in patient demographics \cite{musa2024analyzing}. Population shift is particularly relevant in this context, since factors such as age, disease prevalence, and other patient characteristics can affect both the appearance of medical images and the distribution of the corresponding findings. This makes cross-population evaluation important when assessing whether a model trained on one patient group can generalize to another.

\begin{table}[t]
\centering
\caption{\textbf{Label harmonization between VinDr-CXR
\cite{vindr} and PediCXR \cite{pedicxr}.}
Labels without a corresponding category in the other
dataset are excluded from the five-label target space.}
\label{tab:label_mapping}

\small
\setlength{\tabcolsep}{5pt}
\renewcommand{\arraystretch}{1.15}

\begin{tabular}{@{}lll@{}}
\toprule
\textbf{Target label} &
\textbf{VinDr-CXR} &
\textbf{PediCXR} \\
\midrule
No Finding     & No finding     & No finding \\
Other Diseases & Other diseases & Other diseases \\
Lung Tumor     & Lung tumor     & Lung tumor \\
Pneumonia      & Pneumonia      &
\makecell[l]{
Pneumonia \\
Broncho-pneumonia \\
Pleuro-pneumonia
} \\
Tuberculosis   & Tuberculosis   & Tuberculosis \\
\midrule
\textit{Excluded} & COPD & Remaining global labels \\
\bottomrule
\end{tabular}

\vspace{-0.2cm}
\end{table}

\subsection{Adult-to-Pediatric Domain Shift}


Age-related domain shift is a relevant issue in chest X-ray analysis, particularly given the limited availability of large pediatric datasets \cite{padash2022pediatric}. Rollan-Martinez-Herrera \textit{et al.} \cite{rollan2024cnns} reported a clear drop in pneumonia classification performance when a CNN trained on adult data was evaluated in children, with AUC decreasing from 0.95 to 0.82. Tang \textit{et al.} \cite{tang2019tuna} explored unsupervised domain adaptation and improved pediatric AUC from 0.893, without adaptation, to 0.963 using unlabeled pediatric images. Similar transfer issues have also been studied in lung segmentation \cite{rajaraman2023can,kirscher2025psat}, where performance was shown to vary with pediatric age and anatomy. Capellan-Martin \textit{et al.} \cite{capellan2024zero} further investigated this problem using self-supervised ViT pretraining for zero-shot pediatric tuberculosis detection. Previous studies have mainly considered task-specific CNNs, individual architectures, or single adaptation settings. Recent work has evaluated foundation models across adult and pediatric CXR populations \cite{xu2025benchmarking, fay2025beyond}. However, the effect of downstream adaptation depth has not been systematically compared across multiple foundation-model visual encoders under the same adult-to-pediatric transfer protocol. To the best of our knowledge, no systematic comparison of multiple foundation models with different pretraining domains and objectives under the same adult-to-pediatric CXR evaluation setting has been reported.

%% file: 3_benchmark.tex
\section{Adult-to-Pediatric CXR Benchmark}
\label{sec:bench}

\subsection{Problem Formulation}



\begin{figure*}[t]
    \centering
    \includegraphics[width=1\linewidth]{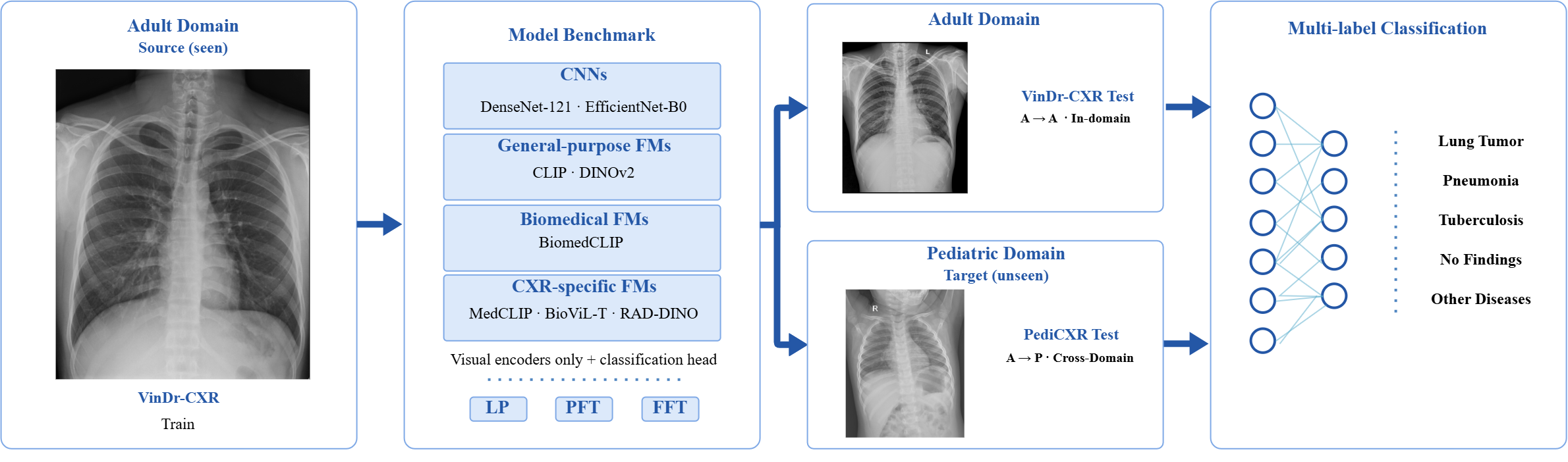}
    \caption{\textbf{Overview of the proposed domain shift evaluation methodology.} Models are trained or adapted using chest X-rays from the seen adult source domain and subsequently evaluated on the unseen pediatric target domain, without target-domain training.}
    \label{fig:benchmark}
\vspace{-0.3cm}
\end{figure*}

We consider multi-label chest X-ray classification under an
adult-to-pediatric domain shift. Let $\mathcal{D}_S$ and
$\mathcal{D}_T$ denote the adult source and pediatric target domains,
respectively. After label harmonization, both domains share a common
set of $C$ clinical findings, with labels $y \in \{0,1\}^{C}$.
Samples from the two domains follow distributions $P_S(X,Y)$ and
$P_T(X,Y)$, where $P_S(X,Y) \neq P_T(X,Y)$.
For each adaptation strategy, a model $f_{\theta}$ is optimized using
only the adult source training set, $\mathcal{D}_S^{\mathrm{train}}$,
and evaluated on both the adult source test set,
$\mathcal{D}_S^{\mathrm{test}}$, and the pediatric target test set,
$\mathcal{D}_T^{\mathrm{test}}$. No samples from the pediatric target
dataset are used during downstream adaptation or model selection. Fig. \ref{fig:benchmark} shows an overview of the evaluation protocol.




\subsection{Datasets}

We use two chest X-ray datasets to study the adult-to-pediatric domain shift: VinDr-CXR \cite{vindr} for the adult population and PediCXR \cite{pedicxr} for the pediatric population. PediCXR \cite{pedicxr} contains 9,125 pediatric chest X-rays, with 7,728 images in the training set and 1,397 in the test set, annotated with 15 global diagnostic labels. Each image has a multi-label, image-level annotation produced by a single assigned radiologist.
VinDr-CXR \cite{vindr} contains 18,000 adult chest X-rays, divided into 15,000 training and 3,000 test images, and provides six image-level labels: Lung Tumor, Pneumonia, Tuberculosis, Other Diseases, COPD, and No Finding. Before constructing the benchmark splits, we performed age-based population screening using the available DICOM metadata. Further details are provided in the Supplementary Material. Since the two datasets use different annotation vocabularies, we harmonize their labels into a common five-label target space consisting of No Finding, Other Diseases, Lung Tumor, Pneumonia, and Tuberculosis. In PediCXR \cite{pedicxr}, Pneumonia, Broncho-pneumonia, and Pleuro-pneumonia are merged into a single Pneumonia label. COPD is excluded from VinDr-CXR \cite{vindr} because no corresponding label is available in PediCXR \cite{pedicxr}, while the remaining PediCXR \cite{pedicxr} labels without a counterpart in VinDr-CXR \cite{vindr} are not considered as target labels. Table~\ref{tab:label_mapping} summarizes this mapping. For the VinDr-CXR \cite{vindr} training set, each image is independently annotated by three radiologists. Their annotations are aggregated into a single multi-label target vector using an element-wise OR operation, implemented as the maximum across the three binary annotation vectors. Thus, a pathology is considered present if it is reported by at least one radiologist, allowing multiple findings from different annotators to be retained simultaneously. To ensure label consistency, No Finding is set to zero whenever any pathological finding is present. In contrast, no annotator aggregation is required for PediCXR \cite{pedicxr}, since each image has a single annotation. This label harmonization procedure is applied consistently to both the training and test splits of each dataset. The distribution of positive samples across the five target labels is reported in Table 2 of the Supplementary Material.

\subsection{Models}

We select a diverse set of models to investigate how different pretraining strategies and domains relate to robustness under the adult-to-pediatric chest X-ray domain shift. Our benchmark spans two main dimensions: the learning paradigm used during pretraining, ranging from supervised learning to image-only self-supervision and image-text contrastive learning, and the domain of the pretraining data, ranging from natural images to biomedical and radiological data. We first consider DenseNet-121 \cite{huang2017densenet} and EfficientNet-B0 \cite{tan2019efficientnet} as conventional CNN baselines. Both models are initialized with weights obtained through supervised pretraining on natural images and provide a reference for architectures that do not rely on large-scale foundation-model pretraining. We then include DINOv2 \cite{oquab2023dinov2} and CLIP \cite{radford2021learning} as general-purpose foundation models pretrained on natural-image data. DINOv2 \cite{oquab2023dinov2} relies on image-only self-supervised learning, whereas CLIP \cite{radford2021learning} uses image-text contrastive pretraining, representing two widely adopted foundation-model pretraining paradigms. To extend the comparison toward domain-specific pretraining, we include BiomedCLIP \cite{zhang2023biomedclip}, which is pretrained on biomedical image-text pairs. Finally, we consider models whose pretraining is more closely aligned with radiological data: RAD-DINO \cite{perez2025rad-dino}, based on image-only self-supervised learning, and BioViL-T \cite{bannur2023biovilt} and MedCLIP \cite{wang2022medclip}, which use image-text pretraining. For all models, we use only the visual encoder and attach the same downstream classification framework. For multimodal models, we do not use the text encoder during downstream training or inference. This common setup ensures that the comparison focuses on the visual representations acquired during pretraining rather than on differences in downstream classification strategies. In this way, these models provide representative points across different pretraining paradigms and levels of domain specificity. Rather than isolating either factor independently, our benchmark lets us assess whether models pretrained with different learning objectives and increasingly domain-relevant data show different levels of robustness when transferring from adult to pediatric chest X-rays.
\vspace{-0.4cm}
\subsection{Adaptation Protocols}


\vspace{-0.1cm}
To evaluate how different levels of model adaptation affect performance under the adult-to-pediatric domain shift, we consider three adaptation protocols: linear probing (LP), partial fine-tuning (PFT), and full fine-tuning (FFT). These protocols progressively increase the number of trainable parameters, ranging from adapting only the classification head to updating the entire visual encoder, allowing us to assess how much modification of the pretrained representations is required for the downstream task \cite{kumar2202fine, Kornblith_2019_CVPR}. In linear probing, the visual encoder is kept frozen, and only the classification head is optimized. This setting evaluates the transferability of the pretrained visual representations without modifying the encoder. In partial fine-tuning, the final encoder block or stage, together with the classification head, is optimized, while the remaining encoder parameters are kept frozen. This provides an intermediate level of adaptation in which high-level representations can adjust to the downstream task while preserving most of the pretrained representation. PFT is defined structurally by unfreezing the final encoder block or stage and is therefore not parameter-matched across architectures. Finally, in full fine-tuning, the entire visual encoder and classification head are jointly optimized, providing the highest degree of adaptation to the training data. Table~\ref{tab:adaptation_parameters} reports the total number of parameters for each model, together with the number of trainable parameters under LP and PFT. Under FFT, all model parameters are trainable by definition.
\vspace{-0.2cm}

\begin{table}[t]
\centering
\caption{\textbf{Model configurations across adaptation strategies.}
Model architectures, total parameter counts, and trainable parameters under linear probing (LP) and partial fine-tuning (PFT). Under full fine-tuning, all model parameters are updated.}
\label{tab:adaptation_parameters}

\scriptsize
\setlength{\tabcolsep}{2.5pt}

\resizebox{\columnwidth}{!}{%
\begin{tabular}{lccccc}
\toprule
\textbf{Model} &
\makecell{\textbf{Visual}\\\textbf{Backbone}} &
\makecell{\textbf{Total}\\\textbf{params.}} &
\makecell{\textbf{LP}\\\textbf{trainable}} &
\makecell{\textbf{PFT}\\\textbf{trainable}} \\
\midrule

DenseNet-121 \cite{huang2017densenet}    & --           & 6.95M  & 5.1K & 2.17M \\
EfficientNet-B0 \cite{tan2019efficientnet} & --           & 4.01M  & 6.4K & 1.14M \\
CLIP \cite{radford2021learning}            & ViT-B/16     & 86.20M & 2.6K & 7.49M \\
DINOv2 \cite{oquab2023dinov2}         & ViT-B/14     & 86.58M & 3.8K & 7.09M \\
BiomedCLIP \cite{zhang2023biomedclip}      & ViT-B/16-224 & 86.19M & 2.6K & 7.49M \\
MedCLIP \cite{wang2022medclip}         & ViT-Swin-tiny& 27.52M & 3.8K & 14.19M \\
BioViL-T \cite{bannur2023biovilt}       & ResNet-50    & 27.27M & 2.6K & 15.49M \\
RAD-DINO \cite{perez2025rad-dino}       & ViT-B/14     & 86.58M & 3.8K & 7.09M \\

\bottomrule
\end{tabular}%
}
\vspace{-0.6 cm}
\end{table}

%% file: 4_evaluation.tex
\section{Evaluation}
\label{sec:eval}

\subsection{In-Domain \& Cross-Domain}
We evaluate all models under two settings: in-domain and cross-domain evaluation. For each adaptation protocol, models are trained exclusively on the VinDr-CXR \cite{vindr} adult training set. In-domain performance is measured on the held-out VinDr-CXR \cite{vindr}  test set, which follows the same source domain as the training data. Cross-domain performance is evaluated on PediCXR \cite{pedicxr}, without any additional training or adaptation using pediatric data. This evaluation protocol allows us to assess not only the predictive performance of each model on the source domain, but also how well it generalizes when transferred to a pediatric population. We compare the in-domain and cross-domain results to quantify the performance degradation associated with the adult-to-pediatric domain shift. The same evaluation procedure is applied to all CNN baselines and foundation models across the three adaptation strategies.
\vspace{-0.2cm}

\subsection{Metrics}

We evaluate performance using Macro-AUROC, mean Average Precision (mAP), and
Macro-F1, computed over the five target labels. Macro-AUROC and mAP are
obtained by independently computing AUROC and AP for each label and taking
their unweighted mean. Macro-F1 is computed analogously using a fixed threshold of 0.5 across all models and labels to ensure a consistent evaluation protocol.  These metrics provide
complementary threshold-independent and threshold-dependent evaluations of
the multi-label classification performance. Macro-AUROC is used as the primary metric for the domain-shift analysis, while mAP and Macro-F1 are reported as complementary metrics. To quantify the degradation under the adult-to-pediatric domain shift, we
additionally report
\begin{equation}
\Delta_{\mathrm{AUROC}}
=
\mathrm{AUROC}_{A \rightarrow A}
-
\mathrm{AUROC}_{A \rightarrow P},
\end{equation}
where $A \rightarrow A$ denotes in-domain adult evaluation and
$A \rightarrow P$ denotes evaluation on the pediatric domain after training
on adult data. Lower $\Delta_{\mathrm{AUROC}}$ indicates a smaller performance
degradation under the domain shift. We distinguish cross-domain performance from robustness to the shift. Cross-domain performance is measured directly by the pediatric AUROC, whereas robustness reflects how much performance is retained relative to the adult in-domain setting and is summarized by $\Delta_{\mathrm{AUROC}}$. We therefore consider both quantities jointly, since a small domain-shift gap alone does not necessarily imply strong performance on the pediatric target domain.

\subsection{Implementation Details}

All experiments were implemented in PyTorch~\cite{paszke2019pytorch} and conducted on an NVIDIA RTX PRO 6000 Blackwell Max-Q Workstation Edition GPU. All input images were resized to 224 × 224 pixels, while retaining the model-specific normalization used by each pretrained encoder. No additional data augmentation was applied during training. Models were trained under the three adaptation protocols described above: linear probing, partial fine-tuning, and full fine-tuning. Following \cite{yang2023rethinking}, the classification head was optimized with a learning rate of $10^{-3}$ in all experiments. For protocols in which the encoder was updated, a learning rate of $10^{-4}$ was used for CNN-based backbones and $10^{-5}$ for Vision Transformer-based backbones. AdamW was used as the optimizer with a weight decay of $10^{-4}$. A learning-rate scheduler was applied to all trainable parameter groups, multiplying the learning rate by $0.7$ every 15 epochs. After age-based screening, a fixed 80/20 split resulted in 11,894 training and 2,974 validation images. Supplementary Tables report detailed split statistics. This split was generated once and kept fixed across all models and adaptation protocols to ensure a consistent comparison. The official VinDr-CXR test set was kept separate and used only for in-domain evaluation. All models were trained for 100 epochs with a batch size of 32. Model selection was performed on the source-domain validation set, with the checkpoint achieving the highest macro-averaged AUROC retained for evaluation. Binary cross-entropy with logits was used as the training objective. To account for class imbalance, class-specific positive weights were computed from the adult training set as the ratio of negative to positive samples for each class and incorporated into the loss. Each experiment was repeated using three different random seeds, and results are reported across the three runs.

\vspace{-0.4cm}
\paragraph{\textbf{Baselines.}}
We include two conventional CNN architectures as baselines to provide a reference for comparison with the evaluated foundation models. DenseNet-121~\cite{huang2017densenet} uses dense connections between layers, which encourage feature reuse and improve information and gradient propagation throughout the network. EfficientNet~\cite{tan2019efficientnet} follows a compound scaling strategy that jointly balances network depth, width, and input resolution, providing a computationally efficient convolutional architecture. These models allow us to assess whether foundation models provide an advantage over established CNN-based approaches under the same training and evaluation settings.
\vspace{-0.2cm}

%% file: 5_results.tex
\section{Results}
\label{sec:results}
\vspace{-0.1cm}
We structure our analysis around four research questions.
\textbf{RQ1} asks which models transfer most effectively from adult to pediatric data. \textbf{RQ2} examines how pretraining domain and supervision paradigm relate to robustness under this shift. \textbf{RQ3} evaluates how linear probing, partial fine-tuning, and full fine-tuning affect pediatric performance and cross-domain robustness. Finally, \textbf{RQ4} assesses whether in-domain performance reliably reflects cross-domain performance and robustness.

\begin{table*}[t]
\caption{\textbf{Performance under the adult-to-pediatric domain shift ($A \rightarrow P$) and in-domain adult evaluation ($A \rightarrow A$) across the three adaptation protocols.} Performance metrics (\%) are reported as mean $\pm$ standard deviation over three random seeds. $\Delta_{\mathrm{AUROC}}$ is computed from the corresponding unrounded mean AUROC values and is defined as $\Delta_{\mathrm{AUROC}} = \mathrm{AUROC}_{A \rightarrow A} - \mathrm{AUROC}_{A \rightarrow P}$, where lower values indicate a smaller performance degradation under the domain shift. Best and second-best mean values within each adaptation strategy are shown in bold and underlined, respectively.}
\label{tab:main_results}
\resizebox{\textwidth}{!}{%
\begin{tabular}{lccccccc}
\hline
\multicolumn{1}{c}{\multirow{2}{*}{Model}} & \multicolumn{3}{c}{\textbf{Cross-Domain (A $\rightarrow$ P)}}                     & \multicolumn{3}{c}{\textbf{In-Domain (A $\rightarrow$ A)}}                        & \multicolumn{1}{l}{\multirow{2}{*}{$\Delta_{\mathrm{AUROC}}\downarrow$}} \\ \cline{2-7}
\multicolumn{1}{c}{}                       & \textbf{Macro AUROC}      & \textbf{mAP}              & \textbf{Macro F1}         & \textbf{Macro AUROC}      & \textbf{mAP}              & \textbf{Macro F1}         & \multicolumn{1}{l}{}                                                     \\ \hline
\multicolumn{8}{c}{\textbf{Linear Probing}}                                                                                                                                                                                                                                                   \\ \hline
DenseNet-121 \cite{huang2017densenet}      & 56.11 \std{0.77}          & 17.63 \std{0.11}          & 15.38 \std{0.50}          & 77.60 \std{0.09}          & 41.58 \std{0.04}          & 38.24 \std{1.02}          & 21.49                                                                    \\
EfficientNet-B0 \cite{tan2019efficientnet} & 46.84 \std{2.17}          & 18.42 \std{0.21}          & 20.47 \std{0.47}          & 74.40 \std{0.51}          & 36.81 \std{0.34}          & 37.91 \std{0.26}          & 27.56                                                                    \\
CLIP \cite{radford2021learning}            & 52.98 \std{0.19}          & 18.29 \std{0.01}          & 21.07 \std{0.26}          & 73.23 \std{0.01}          & 37.43 \std{0.04}          & 38.94 \std{0.16}          & 20.26                                                                    \\
DINOv2 \cite{oquab2023dinov2}              & 57.15 \std{0.11}          & 20.27 \std{0.07}          & 22.09 \std{0.74}          & 83.39 \std{0.02}          & 47.31 \std{0.06}          & 45.21 \std{0.59}          & 26.23                                                                    \\
BiomedCLIP \cite{zhang2023biomedclip}      & 74.95 \std{0.41}          & 20.96 \std{0.92}          & 21.24 \std{1.71}          & 85.43 \std{0.15}          & 52.69 \std{0.68}          & 47.56 \std{3.43}          & \textbf{10.48}                                                           \\
MedCLIP \cite{wang2022medclip}             & \textbf{79.31 \std{0.26}} & \textbf{35.31 \std{0.56}} & \textbf{25.75 \std{0.16}} & {\ul 91.99 \std{0.05}}    & {\ul 64.00 \std{0.25}}    & \textbf{59.02 \std{0.52}} & {\ul 12.69}                                                              \\
BioViL-T \cite{bannur2023biovilt}          & 39.23 \std{0.10}          & 17.39 \std{0.01}          & 13.67 \std{0.04}          & 80.66 \std{0.01}          & 42.98 \std{0.00}          & 41.98 \std{0.03}          & 41.43                                                                    \\
RAD-DINO \cite{perez2025rad-dino}          & {\ul 79.19 \std{0.31}}    & {\ul 24.59 \std{0.22}}    & {\ul 23.98 \std{0.17}}    & \textbf{92.26 \std{0.06}} & \textbf{64.09 \std{0.17}} & {\ul 56.50 \std{0.48}}    & 13.07                                                                    \\ \hline
\multicolumn{8}{c}{\textbf{Partial Fine-Tuning}}                                                                                                                                                                                                                                              \\ \hline
DenseNet-121 \cite{huang2017densenet}      & 63.05 \std{4.84}          & 19.47 \std{0.42}          & 16.18 \std{0.21}          & 81.41 \std{1.69}          & 48.42 \std{2.72}          & 44.01 \std{1.38}          & 18.36                                                                    \\
EfficientNet-B0 \cite{tan2019efficientnet} & 50.36 \std{10.52}         & 19.10 \std{0.38}          & 20.41 \std{1.07}          & 76.38 \std{1.07}          & 39.73 \std{1.09}          & 40.17 \std{0.14}          & 26.02                                                                    \\
CLIP \cite{radford2021learning}            & 50.67 \std{4.04}          & 19.17 \std{0.23}          & 19.78 \std{1.45}          & 72.51 \std{1.47}          & 38.10 \std{0.89}          & 39.46 \std{1.07}          & 21.83                                                                    \\
DINOv2 \cite{oquab2023dinov2}              & 73.44 \std{1.58}          & 23.15 \std{0.29}          & 24.31 \std{0.72}          & 87.25 \std{0.09}          & 57.35 \std{0.43}          & 53.84 \std{1.47}          & 13.81                                                                    \\
BiomedCLIP \cite{zhang2023biomedclip}      & 74.32 \std{1.00}          & \textbf{31.37 \std{0.30}} & 24.23 \std{0.77}          & 86.29 \std{0.44}          & 55.31 \std{0.43}          & 51.30 \std{2.45}          & \textbf{11.97}                                                           \\
MedCLIP \cite{wang2022medclip}             & {\ul 76.81 \std{1.02}}    & {\ul 26.65 \std{0.89}}    & \textbf{26.83 \std{0.73}} & {\ul 91.79 \std{0.12}}    & \textbf{66.61 \std{0.66}} & \textbf{64.03 \std{0.71}} & 14.98                                                                    \\
BioViL-T \cite{bannur2023biovilt}          & 58.34 \std{1.00}          & 20.98 \std{2.30}          & 19.60 \std{1.48}          & 85.80 \std{0.50}          & 54.63 \std{1.41}          & 52.73 \std{3.07}          & 27.46                                                                    \\
RAD-DINO \cite{perez2025rad-dino}          & \textbf{79.76 \std{0.08}} & 26.37 \std{0.45}          & {\ul 25.72 \std{0.24}}    & \textbf{92.42 \std{0.05}} & {\ul 66.11 \std{0.27}}    & {\ul 59.39 \std{1.11}}    & {\ul 12.66}                                                              \\ \hline
\multicolumn{8}{c}{\textbf{Full Fine-Tuning}}                                                                                                                                                                                                                                                 \\ \hline
DenseNet-121 \cite{huang2017densenet}      & 66.35 \std{7.92}          & 20.89 \std{0.56}          & 21.16 \std{0.93}          & 88.35 \std{0.51}          & 58.70 \std{0.95}          & 54.78 \std{4.08}          & 22.00                                                                    \\
EfficientNet-B0 \cite{tan2019efficientnet} & 50.82 \std{6.17}          & 18.39 \std{0.33}          & 16.32 \std{2.56}          & 75.44 \std{2.21}          & 40.81 \std{4.45}          & 35.18 \std{4.78}          & 24.62                                                                    \\
CLIP \cite{radford2021learning}            & 68.52 \std{1.02}          & 20.27 \std{0.79}          & 21.26 \std{1.09}          & 81.70 \std{1.74}          & 50.16 \std{4.08}          & 49.66 \std{3.88}          & \textbf{13.18}                                                           \\
DINOv2 \cite{oquab2023dinov2}              & 71.12 \std{1.07}          & 23.17 \std{2.34}          & 21.31 \std{1.90}          & 84.71 \std{2.11}          & 56.91 \std{3.00}          & 55.05 \std{1.81}          & {\ul 13.59}                                                              \\
BiomedCLIP \cite{zhang2023biomedclip}      & 71.75 \std{1.97}          & 21.91 \std{0.63}          & 21.79 \std{0.56}          & 86.04 \std{1.28}          & 57.41 \std{1.28}          & 55.82 \std{2.05}          & 14.29                                                                    \\
MedCLIP \cite{wang2022medclip}             & {\ul 76.54 \std{1.33}}    & {\ul 25.78 \std{0.92}}    & \textbf{24.56 \std{0.63}} & {\ul 92.39 \std{0.40}}    & \textbf{70.31 \std{0.44}} & \textbf{65.01 \std{3.25}} & 15.85                                                                    \\
BioViL-T \cite{bannur2023biovilt}          & 68.16 \std{4.94}          & 21.00 \std{0.18}          & 22.98 \std{1.11}          & 87.90 \std{0.18}          & 60.94 \std{0.82}          & 60.17 \std{0.81}          & 19.74                                                                    \\
RAD-DINO \cite{perez2025rad-dino}          & \textbf{78.77 \std{0.83}} & \textbf{30.97 \std{5.00}} & {\ul 24.53 \std{0.62}}    & \textbf{93.15 \std{0.41}} & {\ul 68.62 \std{0.28}}    & {\ul 64.92 \std{2.40}}    & 14.38                                                                    \\ \hline
\end{tabular}%
}
\end{table*}

\subsection{Foundation Models vs. CNNs}
RQ1 compares the transfer of ImageNet-pretrained CNNs and foundation models from adult to pediatric domains. As shown in Table~\ref{tab:main_results}, all models degrade under the domain shift, but the magnitude of the drop varies considerably. Under linear probing, the strongest foundation models clearly outperform DenseNet-121 \cite{huang2017densenet} and EfficientNet-B0 \cite{tan2019efficientnet} on the pediatric set and exhibit smaller domain-shift gaps. BiomedCLIP \cite{zhang2023biomedclip}, MedCLIP \cite{wang2022medclip}, and RAD-DINO \cite{perez2025rad-dino} achieve pediatric AUROCs approximately 75\% or higher, compared with 56.11\% and 46.84\% for DenseNet-121  \cite{huang2017densenet} and EfficientNet-B0 \cite{tan2019efficientnet}. This advantage is not universal: DINOv2 \cite{oquab2023dinov2} shows a similar gap to the CNN baselines, while BioViL-T \cite{bannur2023biovilt} suffers the largest drop under linear probing. The advantage changes with stronger adaptation. In particular, DenseNet-121\cite{huang2017densenet} becomes competitive with several foundation models under full fine-tuning, whereas EfficientNet-B0 \cite{tan2019efficientnet} remains less robust. Overall, the strongest foundation models transfer better than the ImageNet-pretrained CNNs, especially under limited adaptation, although foundation-model pretraining alone does not guarantee robustness.

\subsection{Influence of Pretraining on Domain-Shift}
RQ2 compares general-purpose foundation models (CLIP \cite{radford2021learning} and DINOv2 \cite{oquab2023dinov2}), medical vision-language models (BiomedCLIP \cite{zhang2023biomedclip}, MedCLIP \cite{wang2022medclip}, and BioViL-T \cite{bannur2023biovilt}), and the CXR-specific image-only RAD-DINO \cite{perez2025rad-dino}. Clear differences appear across these pretraining settings. RAD-DINO \cite{perez2025rad-dino} shows the most consistent transfer, maintaining a pediatric AUROC of approximately 80\% across all three adaptation protocols. MedCLIP \cite{wang2022medclip} and BiomedCLIP \cite{zhang2023biomedclip} also perform strongly with linear probing, but their pediatric performance decreases as more parameters are fine-tuned. In contrast, BioViL-T \cite{bannur2023biovilt} benefits substantially from stronger adaptation, improving from 39.23\% with linear probing to 68.16\% with full fine-tuning. General-purpose models are similarly sensitive to the adaptation strategy, with DINOv2 \cite{oquab2023dinov2} performing best under partial fine-tuning and CLIP \cite{radford2021learning} under full fine-tuning. Among the evaluated models, the CXR-specific image-only RAD-DINO \cite{perez2025rad-dino} shows the most stable behavior across adaptation protocols, while general-purpose and medical vision-language models exhibit greater variability. 

\subsection{Adaptation on Domain-shift Robustness}

\begin{figure}[t]
    \centering
    \includegraphics[width=\linewidth]{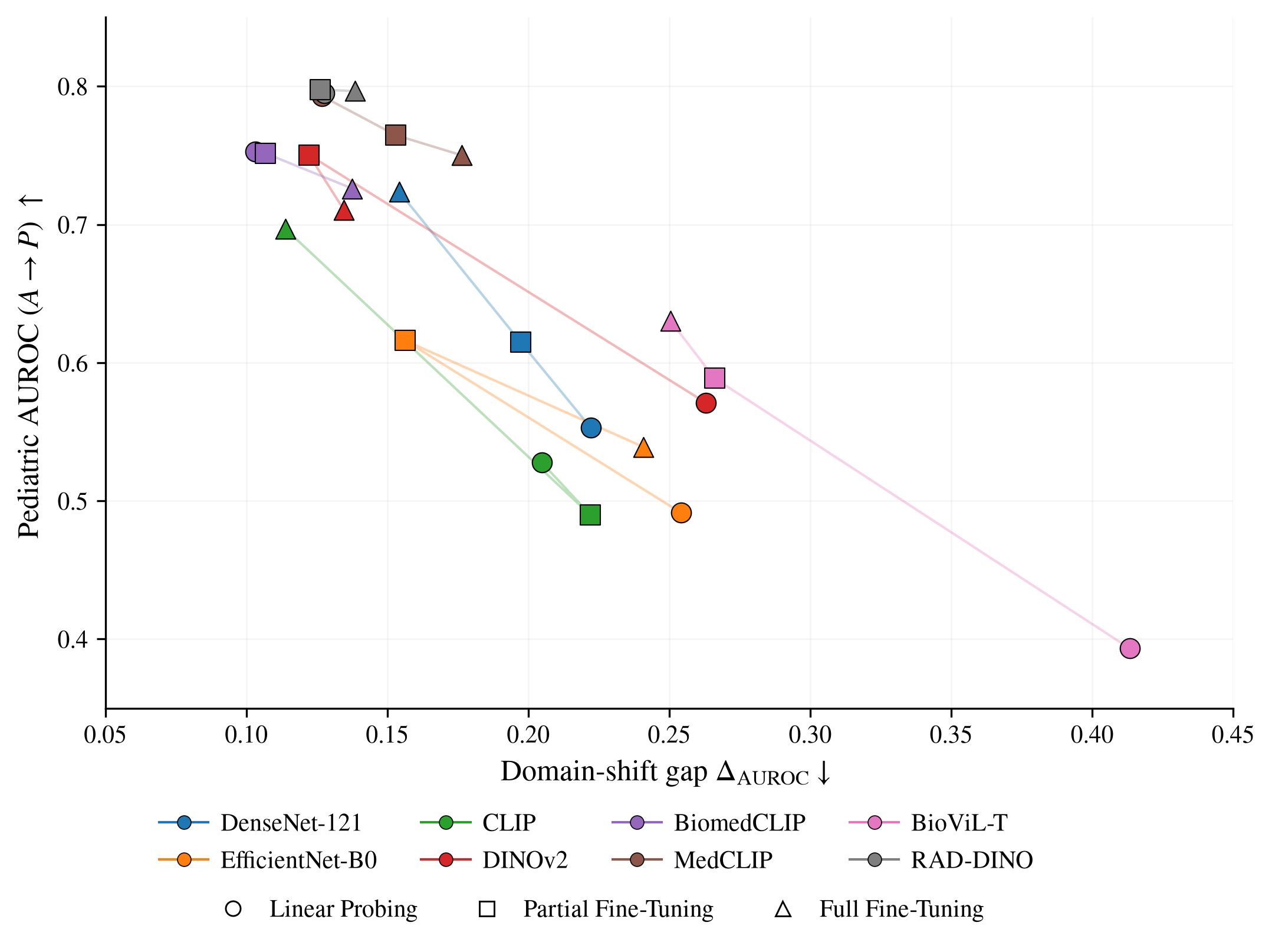}
    \caption{\textbf{Pediatric AUROC vs. Domain-shift gap.} Colors denote models; circles, squares, and triangles denote linear probing, partial fine-tuning, and full fine-tuning, respectively. Lines connect the three protocols for each model. Lower gaps and higher AUROC are better.}
    \label{fig:generalization}
    \vspace{-0.3cm}
\end{figure}

RQ3 examines how linear probing, partial fine-tuning, and full fine-tuning affect pediatric performance and robustness to the adult-to-pediatric shift (Fig.~\ref{fig:generalization}). On average, pediatric AUROC increases from 60.72\% with linear probing to 65.84\% with partial fine-tuning and 69.00\% with full fine-tuning. The corresponding domain-shift gap decreases from 21.65 to 18.39 and 17.21, indicating that most of the reduction is achieved with partial fine-tuning. However, this varies considerably across models. DenseNet-121 \cite{huang2017densenet} and BioViL-T \cite{bannur2023biovilt} benefit from stronger adaptation, whereas DINOv2 performs best with partial fine-tuning. RAD-DINO \cite{perez2025rad-dino} remains almost unchanged across the three protocols, maintaining a pediatric AUROC close to 80\%. In contrast, MedCLIP \cite{wang2022medclip} and BiomedCLIP \cite{zhang2023biomedclip} achieve their best pediatric performance with linear probing and degrade with further fine-tuning. No single adaptation strategy is optimal for all models: additional fine-tuning can improve transfer, have little effect, or reduce cross-domain performance depending on the pretrained representation.

\subsection{In-Domain vs. Cross-Domain Robustness}

\begin{figure*}
    \centering
    \includegraphics[width=1\linewidth]{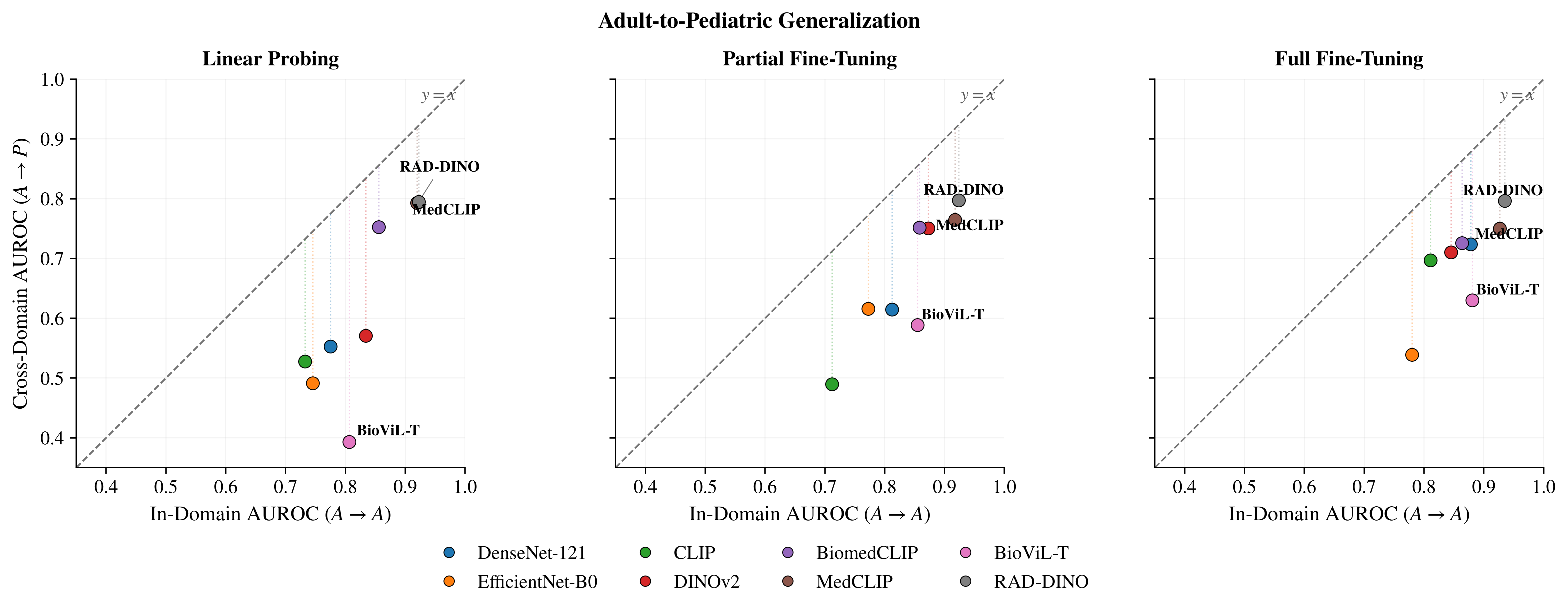}
    \caption{\textbf{Adult in-domain versus pediatric AUROC across adaptation strategies.} The dashed line indicates equal in-domain and cross-domain performance, and the vertical dotted lines show the corresponding domain-shift gap.}
    \label{fig:best}
    \vspace{-0.1cm}
 
\end{figure*}

RQ4 examines whether in-domain performance reliably indicates robustness under the adult-to-pediatric shift. As shown in Fig.~\ref{fig:best}, all configurations lie below the identity line, confirming a systematic drop in pediatric performance. Although stronger in-domain models generally tend to perform better on pediatric data, the size of the domain-shift gap varies substantially. BioViL-T \cite{bannur2023biovilt} illustrates this clearly under linear probing, dropping from 80.66\% in-domain AUROC to 39.23\% on the pediatric set. In contrast, RAD-DINO \cite{perez2025rad-dino} and BiomedCLIP \cite{zhang2023biomedclip} maintain much smaller gaps while also achieving strong pediatric performance. These results show that in-domain performance alone is not sufficient to assess robustness. Models with similar adult performance can behave very differently after transfer to the pediatric domain.

\vspace{-0.3cm}

%% file: 6_discussion.tex
\section{Discussion}
\label{sec:discussion}
\vspace{-0.1cm}
Our results show that foundation models are not uniformly robust to the adult-to-pediatric shift. RAD-DINO \cite{perez2025rad-dino}, MedCLIP \cite{wang2022medclip}, and BiomedCLIP \cite{zhang2023biomedclip} preserve performance relatively well under limited adaptation, whereas BioViL-T \cite{bannur2023biovilt} exhibits the largest domain-shift gap under linear probing. This is notable because BioViL-T \cite{bannur2023biovilt} is also pretrained on medical image-text data. As shown in Table \ref{tab:adaptation_parameters}, it is the only foundation model in our comparison that does not use a ViT-based visual encoder and instead relies on a ResNet-50 backbone. This architectural difference, together with differences in the pretraining objective and data, may partly contribute to its distinct behavior. However, the present experiments do not isolate these factors, so medical-domain pretraining alone cannot be considered sufficient for robustness across adult and pediatric populations. RAD-DINO \cite{perez2025rad-dino} stands out for its stability across adaptation strategies, with pediatric AUROC remaining close to 80\% under linear probing, partial fine-tuning, and full fine-tuning. RAD-DINO and DINOv2 \cite{oquab2023dinov2} provide a particularly informative comparison, as they share the same ViT-B/14 backbone, model size, and adaptation parameter counts. Despite these similarities, DINOv2 shows a substantially larger domain-shift gap under linear probing and benefits more from downstream adaptation. Their different linear-probing behavior is consistent with a benefit from CXR-specific pretraining, although differences in pretraining data and learning objective prevent a causal attribution.
The adaptation experiments also show that more fine-tuning is not always better. Partial fine-tuning captures most of the average reduction in the domain-shift gap observed with full fine-tuning, while the effect of updating additional parameters varies substantially across models. DenseNet-121 \cite{huang2017densenet} and BioViL-T \cite{bannur2023biovilt} benefit from stronger adaptation, whereas RAD-DINO \cite{perez2025rad-dino} changes very little across protocols. MedCLIP \cite{wang2022medclip} and BiomedCLIP \cite{zhang2023biomedclip}, by contrast, achieve their best pediatric performance with linear probing and degrade as more parameters are updated. The choice of adaptation strategy therefore appears to depend strongly on the pretrained representation. Finally, strong in-domain performance does not necessarily imply robustness to the adult-to-pediatric shift. Adult and pediatric performance follow the same general trend, but similar in-domain scores do not always translate into similar behavior after the domain shift. BioViL-T \cite{bannur2023biovilt} is a clear example: despite performing well on the adult test set, its AUROC drops substantially on the pediatric data. This gap shows that in-domain performance alone gives an incomplete view of how well a model transfers across populations. Evaluating models on shifted populations is therefore important when assessing their robustness and transferability.

\vspace{-0.2cm}

%% file: 7_limitations.tex
\section{Limitations}
\label{sec:limitations}
\vspace{-0.3cm}

Although some comparisons are relatively well controlled, most notably RAD-DINO \cite{perez2025rad-dino} and DINOv2 \cite{oquab2023dinov2}, which share the same backbone, model size, and adaptation budget, the full set of models differs in architecture, pretraining data, scale, supervision, and learning objective. Moreover, PFT is defined at the block/stage level and is therefore not parameter-matched across architectures. The observed differences should thus be interpreted as associations rather than as evidence that any single factor, such as image-only supervision or CXR-specific pretraining, is responsible for improved robustness. In addition, our evaluation considers a single adult-to-pediatric cross-dataset domain shift. Because the adult and pediatric populations are drawn primarily from different datasets, age cannot be isolated from dataset-specific differences such as image acquisition. The same trends may therefore not hold across other hospitals, acquisition devices, pediatric populations, or types of distribution shift. The limited number of positive samples for certain pediatric findings also makes class-specific AUROC estimates less reliable, which can affect the interpretation of macro-averaged results. Further evaluation across additional datasets and clinical settings is needed to determine how these findings generalize.
\vspace{-0.5cm}

%% file: 8_conclusion.tex
\section{Conclusion}
\label{sec:conlcusion}
\vspace{-0.3cm}
This is the first work to compare several foundation models under the same adult-to-pediatric setting while also varying the level of model adaptation. We benchmark six representative foundation models and two conventional CNNs for chest X-ray classification, systematically evaluating linear probing, partial fine-tuning, and full fine-tuning across models with different pretraining domains and supervision paradigms. Across this benchmark, all models are affected by the shift from adult to pediatric data, but the extent of the degradation varies substantially across pretrained representations. RAD-DINO \cite{perez2025rad-dino}, MedCLIP \cite{wang2022medclip}, and BiomedCLIP \cite{zhang2023biomedclip} retain comparatively strong pediatric performance with limited adaptation, whereas other models are much more sensitive to the shift. More adaptation is not necessarily better: depending on the model, fine-tuning more of the encoder can improve transfer, have little effect, or even reduce pediatric performance. We also observe that models performing well on the adult test set are not always the ones that transfer best to pediatric data. These results suggest that cross-population robustness depends on both the pretrained representation and how it is adapted to the downstream task. This reinforces the need to evaluate foundation models directly across patient populations, rather than using in-domain performance as a proxy for cross-domain robustness.

%% file: main.bib
@String(CVPR= {IEEE Conf. Comput. Vis. Pattern Recog.})

@String(CVPR  = {CVPR})

@article{vindr,
  title={VinDr-CXR: An open dataset of chest X-rays with radiologist’s annotations},
  author={Nguyen, Ha Q and Lam, Khanh and Le, Linh T and Pham, Hieu H and Tran, Dat Q and Nguyen, Dung B and Le, Dung D and Pham, Chi M and Tong, Hang TT and Dinh, Diep H and others},
  journal={Scientific Data},
  volume={9},
  number={1},
  pages={429},
  year={2022},
  publisher={Nature Publishing Group UK London}
}

@article{pedicxr,
  title={PediCXR: an open, large-scale chest radiograph dataset for interpretation of common thoracic diseases in children},
  author={Pham, Hieu H and Nguyen, Ngoc H and Tran, Thanh T and Nguyen, Tuan NM and Nguyen, Ha Q},
  journal={Scientific data},
  volume={10},
  number={1},
  pages={240},
  year={2023},
  publisher={Nature Publishing Group UK London}
}

@inproceedings{radford2021learning,
  title={Learning transferable visual models from natural language supervision},
  author={Radford, Alec and Kim, Jong Wook and Hallacy, Chris and Ramesh, Aditya and Goh, Gabriel and Agarwal, Sandhini and Sastry, Girish and Askell, Amanda and Mishkin, Pamela and Clark, Jack and others},
  booktitle={International conference on machine learning},
  pages={8748--8763},
  year={2021},
  organization={PmLR}
}

@inproceedings{wang2022medclip,
  title={Medclip: Contrastive learning from unpaired medical images and text},
  author={Wang, Zifeng and Wu, Zhenbang and Agarwal, Dinesh and Sun, Jimeng},
  booktitle={Proceedings of Conference on Empirical Methods in Natural Language Processing},
  pages={3876--3887},
  year={2022}
}

@article{zhang2023biomedclip,
  title={Biomedclip: A multimodal biomedical foundation model pretrained from fifteen million scientific image-text pairs.},
  author={Zhang, Sheng and Xu, Yanbo and Usuyama, Naoto and Xu, Hanwen and Bagga, Jaspreet and Tinn, Robert and Preston, Sam and Rao, Rajesh and Wei, Mu and Valluri, Naveen and others},
  journal={arXiv preprint arXiv:2303.00915},
  year={2023}
}

@article{oquab2023dinov2,
  title   = {{DINOv2}: Learning Robust Visual Features without Supervision},
  author  = {Oquab, Maxime and Darcet, Timoth{\'e}e and Moutakanni, Th{\'e}o and others},
  journal = {Transactions on Machine Learning Research},
  year    = {2024},
  url     = {https://openreview.net/forum?id=a68SUt6zFt}
}

@inproceedings{boecking2022biovil,
  title={Making the most of text semantics to improve biomedical vision--language processing},
  author={Boecking, Benedikt and Usuyama, Naoto and Bannur, Shruthi and Castro, Daniel C and Schwaighofer, Anton and Hyland, Stephanie and Wetscherek, Maria and Naumann, Tristan and Nori, Aditya and Alvarez-Valle, Javier and others},
  booktitle={European Conference on Computer Vision},
  pages={1--21},
  year={2022},
  organization={Springer}
}

@inproceedings{bannur2023biovilt,
  title={Learning to exploit temporal structure for biomedical vision-language processing},
  author={Bannur, Shruthi and Hyland, Stephanie and Liu, Qianchu and Perez-Garcia, Fernando and Ilse, Maximilian and Castro, Daniel C and Boecking, Benedikt and Sharma, Harshita and Bouzid, Kenza and Thieme, Anja and others},
  booktitle={IEEE/CVF Conference on Computer Vision and Pattern Recognition (CVPR)},
  pages={15016--15027},
  year={2023},
  organization={IEEE}
}

@article{tiu2022chexzero,
  title={Expert-level detection of pathologies from unannotated chest X-ray images via self-supervised learning},
  author={Tiu, Ekin and Talius, Ellie and Patel, Pujan and Langlotz, Curtis P and Ng, Andrew Y and Rajpurkar, Pranav},
  journal={Nature Biomedical Engineering},
  volume={6},
  number={12},
  pages={1399--1406},
  year={2022},
  publisher={Nature Publishing Group UK London}
}

@inproceedings{you2023cxr-clip,
  title={Cxr-clip: Toward large scale chest x-ray language-image pre-training},
  author={You, Kihyun and Gu, Jawook and Ham, Jiyeon and Park, Beomhee and Kim, Jiho and Hong, Eun K and Baek, Woonhyuk and Roh, Byungseok},
  booktitle={International Conference on Medical Image Computing and Computer-Assisted Intervention},
  pages={101--111},
  year={2023},
  organization={Springer}
}

@article{xu2023elixr,
  title={Elixr: Towards a general purpose x-ray artificial intelligence system through alignment of large language models and radiology vision encoders},
  author={Xu, Shawn and Yang, Lin and Kelly, Christopher and Sieniek, Marcin and Kohlberger, Timo and Ma, Martin and Weng, Wei-Hung and Kiraly, Atilla and Kazemzadeh, Sahar and Melamed, Zakkai and others},
  journal={arXiv preprint arXiv:2308.01317},
  year={2023}
}

@article{perez2025rad-dino,
  title={Exploring scalable medical image encoders beyond text supervision},
  author={P{\'e}rez-Garc{\'\i}a, Fernando and Sharma, Harshita and Bond-Taylor, Sam and Bouzid, Kenza and Salvatelli, Valentina and Ilse, Maximilian and Bannur, Shruthi and Castro, Daniel C and Schwaighofer, Anton and Lungren, Matthew P and others},
  journal={Nature Machine Intelligence},
  volume={7},
  number={1},
  pages={119--130},
  year={2025},
  publisher={Nature Publishing Group UK London}
}

@article{moutakanni2024raydino,
  title   = {Advancing human-centric AI for robust X-ray analysis through holistic self-supervised learning},
  author  = {Moutakanni, Th{\'e}o and Bojanowski, Piotr and Chassagnon, Guillaume and others},
  journal = {Nature Communications},
  volume  = {17},
  pages   = {9931},
  year    = {2026},
  doi     = {10.1038/s41467-026-76076-4}
}

@article{yao2025evax,
  title={Eva-x: A foundation model for general chest x-ray analysis with self-supervised learning},
  author={Yao, Jingfeng and Wang, Xinggang and Song, Yuehao and Zhao, Huangxuan and Ma, Jun and Chen, Yajie and Liu, Wenyu and Wang, Bo},
  journal={npj Digital Medicine},
  volume={8},
  number={1},
  pages={678},
  year={2025},
  publisher={Nature Publishing Group UK London}
}

@inproceedings{huang2017densenet,
  title={Densely connected convolutional networks},
  author={Huang, Gao and Liu, Zhuang and Van Der Maaten, Laurens and Weinberger, Kilian Q},
  booktitle={IEEE Conference on Computer Vision and Pattern Recognition (CVPR)},
  pages={2261--2269},
  year={2017},
  organization={Ieee}
}

@inproceedings{tan2019efficientnet,
  title={Efficientnet: Rethinking model scaling for convolutional neural networks},
  author={Tan, Mingxing and Le, Quoc},
  booktitle={International Conference on Machine Learning},
  pages={6105--6114},
  year={2019},
  organization={PmLR}
}

@article{padash2022pediatric,
  title={Pediatric chest radiograph interpretation: how far has artificial intelligence come? A systematic literature review},
  author={Padash, Sirwa and Mohebbian, Mohammad Reza and Adams, Scott J and Henderson, Robert DE and Babyn, Paul},
  journal={Pediatric Radiology},
  volume={52},
  number={8},
  pages={1568--1580},
  year={2022},
  publisher={Springer}
}

@article{chatterjee2025children,
  title={Children are not small adults: addressing limited generalizability of an adult deep learning CT organ segmentation model to the pediatric population},
  author={Chatterjee, Devina and Kanhere, Adway and Doo, Florence X and Zhao, Jerry and Chan, Andrew and Welsh, Alexander and Kulkarni, Pranav and Trang, Annie and Parekh, Vishwa S and Yi, Paul H},
  journal={Journal of Imaging Informatics in Medicine},
  volume={38},
  number={3},
  pages={1628--1641},
  year={2025},
  publisher={Springer}
}

@article{cheng2024self,
  title={Self-supervised learning for building robust pediatric chest x-ray classification models},
  author={Cheng, Sheng and Starosolski, Zbigniew A and Subramanian, Devika},
  journal={arXiv preprint arXiv:2409.00231},
  year={2024}
}

@article{zunaed2024improving,
  title={Improving Pediatric Pneumonia Diagnosis with Adult Chest X-ray Images Utilizing Contrastive Learning and Embedding Similarity},
  author={Zunaed, Mohammad and Hasan, Anwarul and Hasan, Taufiq},
  journal={arXiv preprint arXiv:2404.12958},
  year={2024}
}

@article{moreno2012unifying,
  title={A unifying view on dataset shift in classification},
  author={Moreno-Torres, Jose G and Raeder, Troy and Alaiz-Rodr{\'\i}guez, Roc{\'\i}o and Chawla, Nitesh V and Herrera, Francisco},
  journal={Pattern Recognition},
  volume={45},
  number={1},
  pages={521--530},
  year={2012},
  publisher={Elsevier}
}

@article{guan2021domain,
  title={Domain adaptation for medical image analysis: a survey},
  author={Guan, Hao and Liu, Mingxia},
  journal={IEEE Transactions on Biomedical Engineering},
  volume={69},
  number={3},
  pages={1173--1185},
  year={2021},
  publisher={IEEE}
}

@article{bommasani2021opportunities,
  title={On the opportunities and risks of foundation models},
  author={Bommasani, Rishi and Hudson, Drew A and Adeli, Ehsan and Altman, Russ and Arora, Simran and von Arx, Sydney and Bernstein, Michael S and Bohg, Jeannette and Bosselut, Antoine and Brunskill, Emma and others},
  journal={arXiv preprint arXiv:2108.07258},
  year={2021}
}

@article{van2025foundation,
  title={Foundation models in medical imaging: A review and outlook},
  author={van Veldhuizen, Vivien and Botha, Vanessa and Lu, Chunyao and Cesur, Melis Erdal and Lipman, Kevin Groot and de Jong, Edwin D and Horlings, Hugo and Sanchez, Cl{\'a}risa I and Snoek, Cees GM and Wessels, Lodewyk and others},
  journal={arXiv preprint arXiv:2506.09095},
  year={2025}
}

@inproceedings{musa2024analyzing,
  title={Analyzing cross-population domain shift in chest X-ray image classification and mitigating the gap with deep supervised domain adaptation},
  author={Musa, Aminu and Ibrahim Adamu, Mariya and Kakudi, Habeebah Adamu and Hernandez, Monica and Lawal, Yusuf},
  booktitle={International Conference on Medical Image Computing and Computer-Assisted Intervention},
  pages={585--595},
  year={2024},
  organization={Springer}
}

@inproceedings{kirscher2025psat,
  title={PSAT: Pediatric Segmentation Approaches via Adult Augmentations and Transfer Learning},
  author={Kirscher, Tristan and Faisan, Sylvain and Coubez, Xavier and Barrier, Loris and Meyer, Philippe},
  booktitle={International Conference on Medical Image Computing and Computer-Assisted Intervention},
  pages={474--483},
  year={2025},
  organization={Springer}
}

@misc{yang2023rethinking,
      title={Rethinking Foundation Models for Medical Image Classification through a Benchmark Study on MedMNIST}, 
      author={Fuping Wu and Bartlomiej W. Papiez},
      year={2025},
      eprint={2501.14685},
      archivePrefix={arXiv}
}

@InProceedings{Kornblith_2019_CVPR,
author = {Kornblith, Simon and Shlens, Jonathon and Le, Quoc V.},
title = {Do Better ImageNet Models Transfer Better?},
booktitle = {Proceedings of the IEEE/CVF Conference on Computer Vision and Pattern Recognition (CVPR)},
month = {June},
year = {2019}
}

@inproceedings{kumar2202fine,
  title     = {Fine-Tuning Can Distort Pretrained Features and Underperform Out-of-Distribution},
  author    = {Kumar, Ananya and Raghunathan, Aditi and Jones, Robbie and Ma, Tengyu and Liang, Percy},
  booktitle = {International Conference on Learning Representations},
  year      = {2022},
  url       = {https://openreview.net/forum?id=UYneFzXSJWh}
}

@article{paszke2019pytorch,
  title={Pytorch: An imperative style, high-performance deep learning library},
  author={Paszke, Adam and Gross, Sam and Massa, Francisco and Lerer, Adam and Bradbury, James and Chanan, Gregory and Killeen, Trevor and Lin, Zeming and Gimelshein, Natalia and Antiga, Luca and others},
  journal={Advances in neural information processing systems},
  volume={32},
  year={2019}
}

@article{rollan2024cnns,
  title={CNNs trained with adult data are useful in pediatrics. A pneumonia classification example},
  author={Rollan-Martinez-Herrera, Maria and D{\'\i}az, Alejandro A and Est{\'e}par, Rub{\'e}n San Jos{\'e} and Sanchez-Ferrero, Gonzalo Vegas and Ross, James C and Est{\'e}par, Ra{\'u}l San Jos{\'e} and Nardelli, Pietro},
  journal={Plos one},
  volume={19},
  number={7},
  pages={e0306703},
  year={2024},
  publisher={Public Library of Science San Francisco, CA USA}
}

@inproceedings{tang2019tuna,
  title={Tuna-net: Task-oriented unsupervised adversarial network for disease recognition in cross-domain chest x-rays},
  author={Tang, Yuxing and Tang, Youbao and Sandfort, Veit and Xiao, Jing and Summers, Ronald M},
  booktitle={International Conference on Medical Image Computing and Computer-Assisted Intervention},
  pages={431--440},
  year={2019},
  organization={Springer}
}

@inproceedings{capellan2024zero,
  title={Zero-shot pediatric tuberculosis detection in chest X-rays using self-supervised learning},
  author={Capell{\'a}n-Mart{\'\i}n, Daniel and Parida, Abhijeet and G{\'o}mez-Valverde, Juan J and Sanchez-Jacob, Ramon and Roshanitabrizi, Pooneh and Linguraru, Marius G and Ledesma-Carbayo, Mar{\'\i}a J and Anwar, Syed M},
  booktitle={2024 IEEE International Symposium on Biomedical Imaging (ISBI)},
  pages={1--5},
  year={2024},
  organization={IEEE}
}

@article{rajaraman2023can,
  title={Can deep adult lung segmentation models generalize to the pediatric population?},
  author={Rajaraman, Sivaramakrishnan and Yang, Feng and Zamzmi, Ghada and Xue, Zhiyun and Antani, Sameer},
  journal={Expert systems with applications},
  volume={229},
  pages={120531},
  year={2023},
  publisher={Elsevier}
}

@inproceedings{fay2025beyond,
  title={Beyond the prompt: deploying medical foundation models on diverse chest X-ray populations},
  author={Fay, Louisa and Delbrouck, Jean-Benoit and K{\"u}stner, Thomas and Yang, Bin and Codella, Noel C and Lungren, Matthew P and Langlotz, Curtis and Gatidis, Sergios},
  booktitle={Medical Imaging with Deep Learning},
  year={2025}
}

@article{xu2025benchmarking,
  title={Benchmarking Chest X-ray Diagnosis Models Across Multinational Datasets},
  author={Xu, Qinmei and Li, Yiheng and Zhan, Xianghao and Er, Ahmet Gorkem and Dashevsky, Brittany and Xu, Chuanjun and Alawad, Mohammed and Yang, Mengya and Ya, Liu and Zhou, Changsheng and Li, Xiao and Itakura, Haruka and Gevaert, Olivier},
  journal={arXiv preprint arXiv:2505.16027},
  year={2025}
}
